\documentclass[letterpaper]{article} 
\usepackage{aaai25}  
\usepackage{times}  
\usepackage{helvet}  
\usepackage{courier}  
\usepackage[hyphens]{url}  
\usepackage{graphicx} 
\usepackage{natbib}  
\usepackage{caption} 
\usepackage{algorithm}
\usepackage{algorithmic}
\usepackage{amsmath}
\usepackage{amssymb}
\usepackage{enumitem}
\usepackage{multicol}
\usepackage{multirow}
\usepackage{url}
\usepackage{pifont} 

\newcommand{\ie}{\textit{i}.\textit{e}.}

\usepackage{newfloat}
\usepackage{listings}
\DeclareCaptionStyle{ruled}{labelfont=normalfont,labelsep=colon,strut=off} 
\floatstyle{ruled}
\newfloat{listing}{tb}{lst}{}
\floatname{listing}{Listing}
\title{SSLFusion: Scale \& Space Aligned Latent Fusion Model for Multimodal 3D Object Detection}

\author{
    Bonan Ding\textsuperscript{\rm 1},
    Jin Xie\textsuperscript{\rm 1}\footnote{Corresponding Author.},
    Jing Nie\textsuperscript{\rm 2},
    Jiale Cao\textsuperscript{\rm 3,4}
}
\affiliations {
    \textsuperscript{\rm 1}School of Big Data and Software Engineering, Chongqing University, Chongqing, China\\
    \textsuperscript{\rm 2}School of Microelectronics and Communication Engineering, Chongqing University, Chongqing, China\\    
    \textsuperscript{\rm 3}School of Electrical and Information Engineering, Tianjin University, Tianjin, China\\
    \textsuperscript{\rm 4}Shanghai Artificial Intelligence Laboratory, Shanghai, China.\\    
    202224131052@stu.cqu.edu.cn, 
    xiejin@cqu.edu.cn,
    jingnie@cqu.edu.cn,
    connor@tju.edu.cn
}

\begin{document}

\maketitle

\begin{abstract}
Multimodal 3D object detection based on deep neural networks has indeed made significant progress.
However, it still faces challenges due to the misalignment of scale and spatial information between features extracted from 2D images and those derived from 3D point clouds.
Existing methods usually aggregate multimodal features at a single stage. 
However, leveraging multi-stage cross-modal features is crucial for detecting objects of various scales. 
Therefore, these methods often struggle to integrate features across different scales and modalities effectively, thereby restricting the accuracy of detection.
Additionally, the time-consuming Query-Key-Value-based (QKV-based) cross-attention operations often utilized in existing methods aid in reasoning the location and existence of objects by capturing non-local contexts. However, this approach tends to increase computational complexity.
To address these challenges, we present SSLFusion, a novel Scale \& Space Aligned Latent Fusion Model, consisting of a scale-aligned fusion strategy (SAF), a 3D-to-2D space alignment module (SAM), and a latent cross-modal fusion module (LFM). 
SAF mitigates scale misalignment between modalities by aggregating features from both images and point clouds across multiple levels. 
SAM is designed to reduce the inter-modal gap between features from images and point clouds by incorporating 3D coordinate information into 2D image features.
Additionally, LFM captures cross-modal non-local contexts in the latent space without utilizing the QKV-based attention operations, thus mitigating computational complexity. 
Experiments on the KITTI and DENSE datasets demonstrate that our SSLFusion outperforms state-of-the-art methods.
Our approach obtains an absolute gain of 2.15\% in 3D AP, compared with the state-of-art method GraphAlign on the moderate level of the KITTI \textit{test} set. 
\end{abstract}
\begin{links}
    \link{Code}{https://github.com/TLDBN/SSLFusion}
\end{links}
\section{Introduction}

\begin{figure}[ht]
  \centering
  \includegraphics[width=\linewidth]{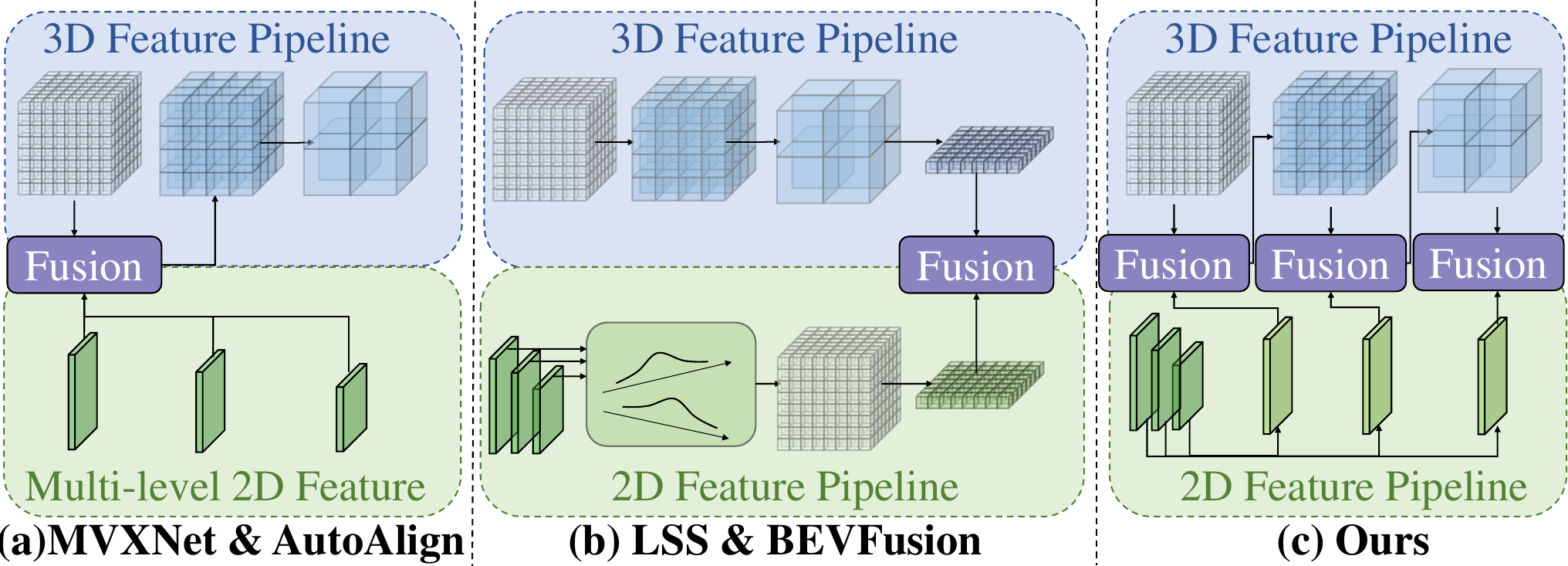}
  \caption{Comparison of our fusion strategy with other methods. Feature-level fusion methods can be divided into two categories: (a) Early-Fusion methods fuse multi-scale image feature maps with the first-stage voxel features. (b) Late-Fusion methods employ depth estimation to transform multi-scale 2D features into 3D space or BEV space, subsequently fusing them with 3D features in those spaces at a single stage. (c) In contrast, our approach fuses multi-stage multi-scale 2D and 3D features in an alignment manner, as opposed to the single-stage fusion in categories (a) and (b).}
  \label{fig:compare}
\end{figure}

\begin{figure*}[t]
  \centering
  \includegraphics[width=\textwidth]{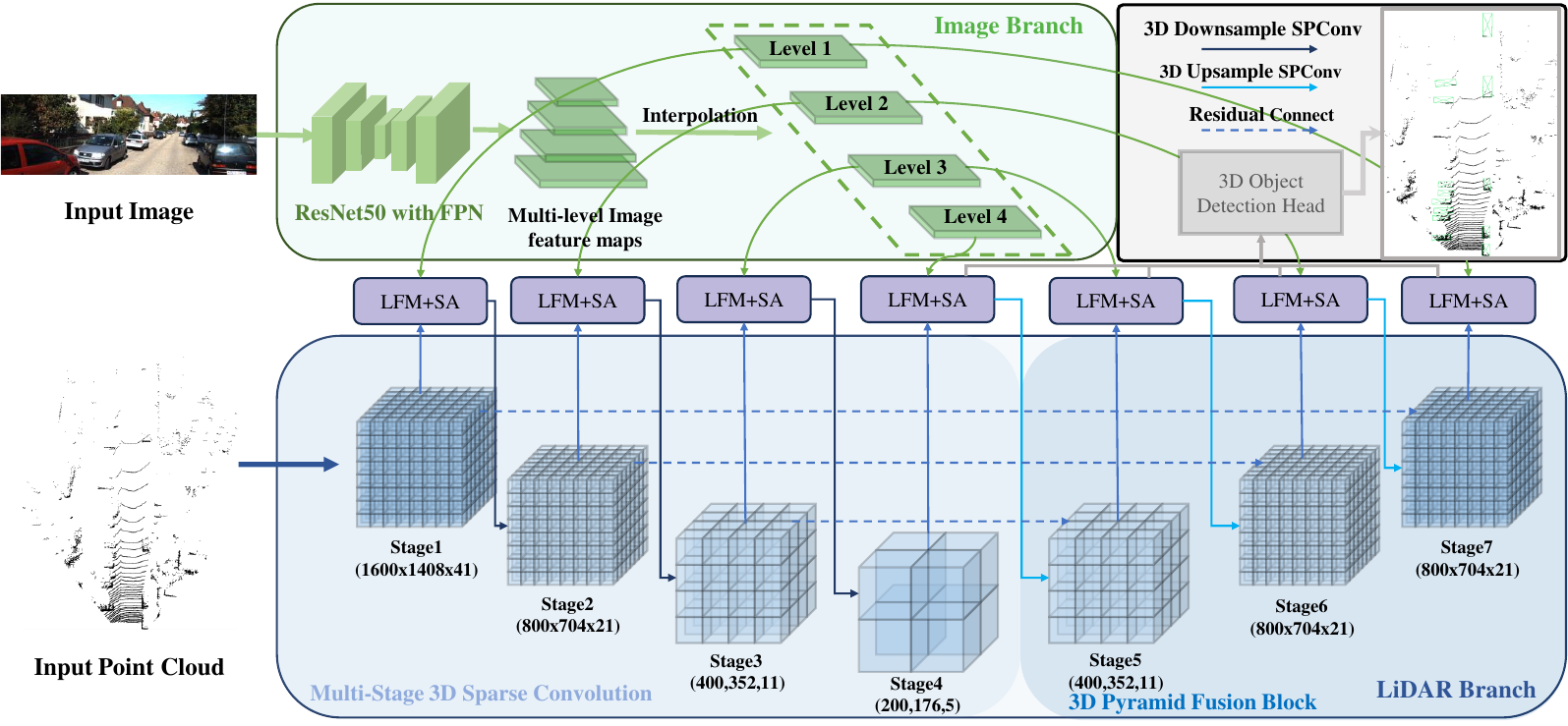}
  \caption{The overall architecture of SSLFusion. Our model consists of four parts: Lidar branch and image branch to extract Lidar and image features, respectively; the proposed Latent Cross-Modal Fusion module with Space Alignment fuses LiDAR and image features at each stage in space aligned manner; and the 3D object detection head generates 3D object detection results based on the multi-level fusion features.}
  \label{fig:overview}
\end{figure*}

The rapid advancement of autonomous vehicles and robotics has driven the need for advanced perception systems capable of accurately interpreting the 3D environment. A critical component of these systems is the ability to detect and classify objects in 3D space, which is essential for safe navigation and decision-making. Multimodal 3D object detection, which combines LiDAR and camera data, has emerged as a promising solution due to the complementary nature of these sensors—point clouds providing precise geometric information and RGB images offering rich semantic context.

Existing feature-level fusion multimodal 3D object detection methods have advanced, but challenges remain. These methods can be categorized into early-fusion, late-fusion, and multi-stage fusion approaches. For example, early-fusion methods like those using matrix projection calibration to combine multi-scale image features with single-scale voxel features~\cite{MVXNet, Graphalign} face difficulties in effectively integrating high-level semantic information from cameras with geometric details from LiDAR. Additionally, this fusion strategy may cause a loss of spatial details when features are downsampled by 3D sparse convolutions, as shown in Fig.~\ref{fig:compare} (a), which is crucial for detecting small objects.

Late-fusion approaches~\cite{BEVFusion, HFMI} process the modalities separately and combine the two types of features in local Regions of Interest (RoI) or Bird's Eye View (BEV) space. These methods often utilize depth estimation to project 2D features into 3D space for fusion. While these methods can leverage the semantic richness of 2D features, they are computationally intensive and can be less accurate due to the inherent uncertainty in depth estimation. This process is also prone to introduce errors when dealing with complex scenes or objects at varying distances.

Both early and late fusion methods struggle to align features across different scales and modalities. The resolution and field-of-view discrepancies between LiDAR and camera data can result in misalignment, particularly in cluttered environments or with objects of varying sizes. Besides, existing feature-level fusion methods rely on global cross-modal interactions through self and cross-modal attention mechanisms~\cite{DeepFusion, BEVFusion, HFMI}, which increases computational complexity. These methods also struggle with the curse of dimensionality when fusing features from both modalities, leading to inefficient resource usage and reduced detection performance.

To address the above limitations, we introduce a novel Scale \& Space Aligned Latent Fusion (SSLFusion) model for multi-stage cross-modality fusion. SSLFusion is designed to align multi-stage features from both modalities in a way that mitigates scale misalignment and information loss. 
Specifically, our model employs a Scale-Aligned Fusion strategy to ensure that features from 2D and 3D branches are fused at each stage to leverage the strengths of both modalities without losing detailed spatial information. 
Additionally, our 3D-to-2D Space Alignment technique effectively bridges the gap between the 3D and 2D spaces, enabling a more accurate and coherent fusion of features. SSLFusion also incorporates a Latent Cross-Modal Fusion module that facilitates efficient interaction between the modalities, further enhancing the quality of the fused features and the subsequent detection performance efficiently.

The efficacy of SSLFusion is demonstrated through extensive experiments on KITTI~\cite{KITTI} and DENSE~\cite{Dense}, showcasing its superior performance across various conditions and object scales. An ablation study further validates the contributions of our model. To summarize, our contributions are as follows: 

\begin{itemize}[itemsep=2pt,topsep=0pt,parsep=0pt]
    \item We propose SSLFusion, a novel multi-stage fusion model that aligns multi-stage features from LiDAR and camera data in both scale and space, addressing the challenges of misalignment and information loss.
    \item We introduce a Scale-Align Fusion strategy that fuses features at each stage of the detection pipeline, ensuring alignment of receptive fields for cross-modal fusion across different scales, while simultaneously reducing fusion information loss during the downsampling process.
    \item Our 3D-to-2D Space Alignment technique provides a new approach to imbue 3D spatial position information into 2D image features, reducing the gap between the two modalities' spatial disparities before cross-modal fusion.
    \item Extensive experimental results on KITTI and DENSE datasets validate the effectiveness of SSLFusion and its components, demonstrating its robustness and generalization capabilities in various driving scenarios.
\end{itemize}

\begin{figure}[t]
  \centering
  \includegraphics[width=\linewidth]{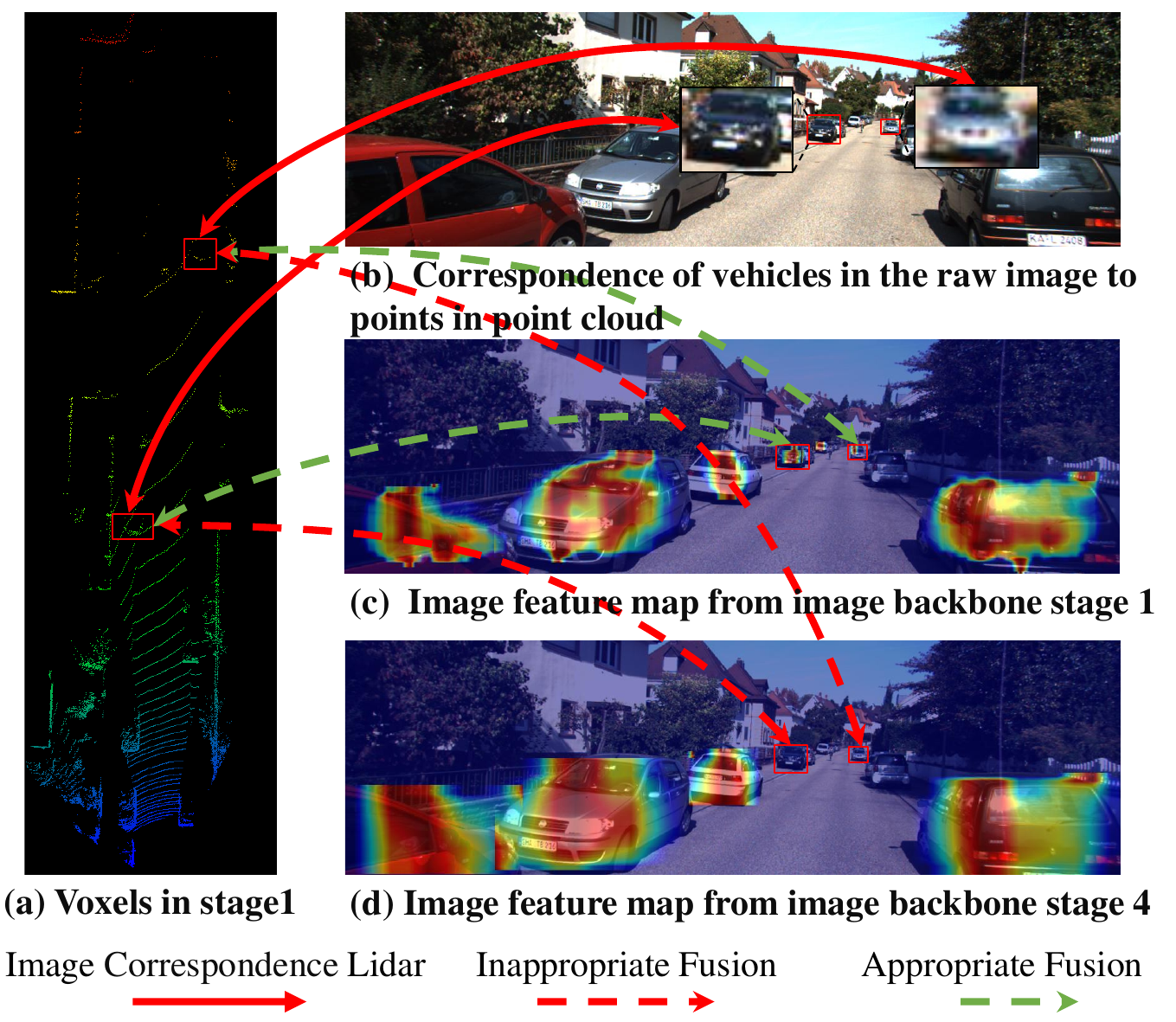}
  \caption{Description of the alignment fusion of different levels of image features and voxels. (a) and (b) demonstrate the alignment relationship between the pixels of distant objects of images and the voxels of the first stage of the 3D backbone. The voxels contained in distant objects are fewer, and their quantity further reduces with the downsampling of 3D convolutions. (c) and (d) depict the feature attention of the first and fourth-level image feature obtained by the image backbone on objects. It can be observed from the figure that the image features of level 1 have foreground features for distant objects, while stage 4 only has background features. Thus, fusing the Stage 1 voxel in 3D with the image features of Stage 4 would introduce noisy features.}
  \label{fig:scale-align}
\end{figure}

\section{Methods}
\subsection{Framework Overview}
\label{subsec:overview}

The overall architecture of SSLFusion is shown in Fig.~\ref{fig:overview} with four components:(a) an image branch with a 2D backbone that extracts multi-level image feature maps, (b) a LiDAR branch containing multi-stage 3D sparse convolutions with 3D pyramid fusion, (c) a latent cross-modal fusion module with space alignment which fuses image features from the image branch and 3D voxel features from each 3D sparse convolution stage in the 3D backbone. (d) a 3D object detection head that generates 3D object locations and classification scores based on the multi-level fusion features from multi-stage latent cross-modal fusion modules.

\subsection{Scale-Aligned Fusion}
\label{subsec:SA3DFP}
Most existing feature-level Lidar-Camera fusion 3D object detection methods ignore or care less about the scale alignment in the fusion stage. As shown in Fig.~\ref{fig:compare}, some early-fusion methods, such as MVX-Net~\cite{MVXNet} and Auto-Alignv2~\cite{AutoAlignv2} utilize matrix projection calibration to fuse multi-level image feature maps with single-stage voxel features. 
While the latter method introduces cross-modality attention feature alignment to alleviate the impact of noisy image features, the detriment remains inevitable. 
Moreover, the early-fusion may lead to loss of image modal information during the multi-stage 3D sparse convolution downsampling process. LSS~\cite{LSS} and BEVFusion~\cite{BEVFusion} are typical late-fusion methods that utilize depth estimation to elevate single or multi-level image features into three-dimensional space based on depth distribution estimation. However, this process is highly time-consuming and resource-intensive in terms of computation. Furthermore, it is constrained by the imprecise nature of depth estimation. To address the aforementioned challenges, we present a novel approach called Scale-Aligned Fusion. Specifically, the scale-aligned fusion strategy is devised to mitigate the issue of scale-misaligned information fusion. 
Additionally, a 3D pyramid fusion block is designed to minimize information loss, boost the semantic content of fused features, and capture details across various scales in 3D space.

\noindent\textbf{Scale-Aligned Fusion Strategy.}
Features are processed using downsampling operations across multiple stages in 2D or 3D convolutional backbones. 
It results in features of each stage containing information from different receptive fields. The goal of the scale-aligned fusion strategy is to sequentially fuse the features from both the 2D and 3D branches at each stage, which aligns the receptive field information and prevents the fusion of incorrect information.

For multi-level image features from the image branch,
deeper-level feature maps have smaller sizes, resulting in each pixel encompassing a larger receptive field. Consequently, these
deeper-level feature maps tend to lack foreground information on small objects and primarily contain background information over a broader range, while 
shallower-level features exhibit the opposite behavior. Similarly, during the downsampling operation of 3D sparse convolution, the number of voxels decreases, akin to the features in the image branch, with fewer voxels in deeper levels and more in shallower levels. This discrepancy is evident in 3D voxel data, where objects further away contain fewer voxels, as depicted in Fig.~\ref{fig:scale-align}, the fusion between the shallow-stage voxel features of 3D sparse convolution and the deep-level image features may result in excessive incorporation of noisy background information. Conversely, performing scale-corresponding fusion would minimize the fusion of erroneous information to the greatest extent possible. 

\noindent\textbf{3D Pyramid Fusion Block.}
Inspired by Feature Pyramid Network~\cite{FPN} and voxel-FPN~\cite{VoxelFPN}, we propose a 3D pyramid fusion block, following the 3D sparse convolution backbone, as shown in Fig.~\ref{fig:overview}. This block processes multi-level fused 3D features from the backbone by performing layer-wise upsampling using sparse convolution weights and establishing lateral connections. These connections add the upsampled 3D features to corresponding backbone levels, effectively fusing image features, reducing information loss, and enhancing multi-scale information. Notably, the upsampling operation in the 3D pyramid fusion block is the inverse of the backbone’s downsampling, meaning it doesn’t increase the model’s parameter count, ensuring high computational efficiency.

\begin{figure}[t]
  \centering
  \includegraphics[width=\linewidth]{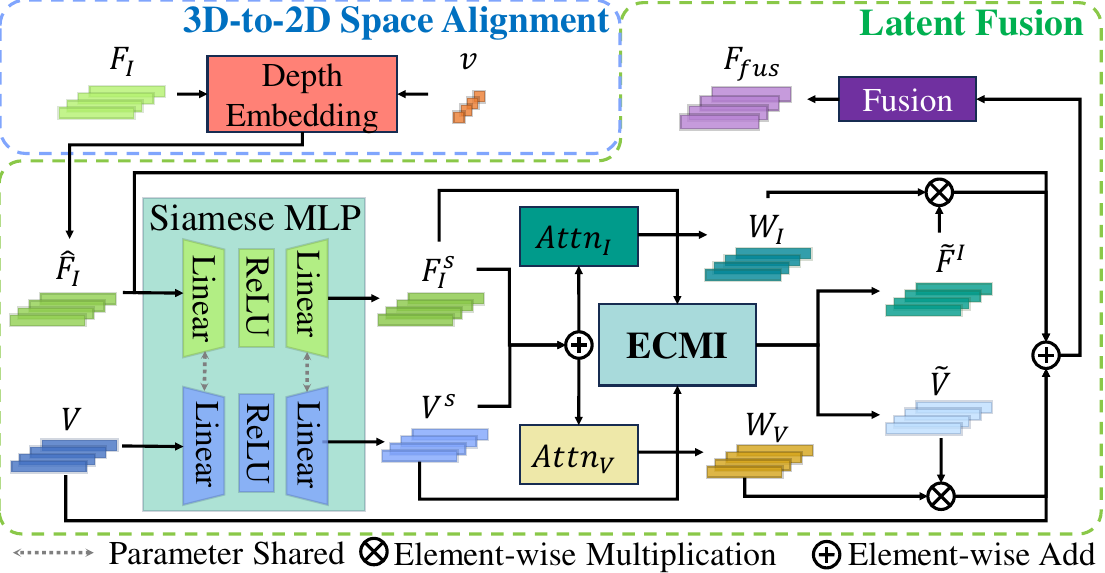}
  \caption{Structure of Latent Cross-Modal Fusion Module with 3D-to-2D Space Alignment.}
  \label{fig:DLGFM}
\end{figure}

\subsection{3D-to-2D Space Alignment}
\label{subsec:spacealign}
Even though we employ the scale-aligned fusion strategy to alleviate the scale misalignment issue in cross-space-modal fusion, some misalignment persists due to the inherent gap between 2D and 3D spaces. 
DeepFusion~\cite{DeepFusion} introduces a technique named Learnable-Align that utilizes a cross-attention mechanism to dynamically capture correlations between two modalities. HFMI~\cite{HFMI} adopts the linear increasing depth discretization (LID) method~\cite{Center3d} to discretize the depth map generated by the point cloud. It produces depth bins used to lift 2D feature maps into 3D space for homogeneous fusion, aiming to mitigate the gap between modalities. 
Inspired by previous methods, we introduce a straightforward yet innovative approach termed 3D-to-2D space alignment, which simply incorporates spatial location information of 3D space into 2D features by encoding the voxel center positions corresponding to image features. By integrating this 3D space information into the image features at each position, we assess them in a trainable manner to alleviate the disparity between the two modalities.
Specifically, in stage $j$, given the voxels' center coordinates $v_i^j = (x_i^j, y_i^j, z_i^j), \forall i \in \{1,...,N_j\}$, calibrate matrix $CM$ and $j~th$ level image feature map $I^j_{fpn} \in \mathbb{R}^{H_j \times W_j \times C}$ from the image FPN. We first upsample the image feature map to the original image size $(H, W)$ by bilinear interpolation, termed as $I^j \in \mathbb{R}^{H \times W \times C}$. Then we can get corresponded sparse image features $F_I^j \in \mathbb{R}^{N_j \times C}$ by $F_I^j = v^j \cdot CM$. The 3D depth-embedding $D$ generated by:
\begin{equation}
    D = sigmoid(fc_2(\alpha(fc_1(v^j)))),
\end{equation}
where $fc_1$ and $fc_2$ represent two fully connected layers, and $\alpha$ denotes a hyperbolic tangent activation function (Tanh). After that, we can get the depth-guided sparse image features $\hat{F}_I^j = D \cdot F_I^j$.\\

\subsection{Latent Cross-Modal Fusion}
\label{subsec:LCF}
Existing deep feature fusion methods~\cite{DeepFusion, HFMI, AutoAlign} use cross-modal attention mechanism to fuse LiDAR and image features, in terms of $Q, K, V$ manner, which could aggregate global cross-modal information better. However, the quadratic complexity of these methods is $O(N^2 \cdot c)$ for $N$ voxels or points with channel dimension $c$. To decrease the space and time complexity in the fusion stage, we propose a Latent Cross-Modal Fusion strategy for effectively fusing the sparse image features with their corresponding voxel features in a non-local manner inspired by LatentGNN~\cite{LatentGNN} and Siamese Representation Learning~\cite{SRL}. 
After obtaining the 3D-to-2d space-aligned sparse image features $\hat{F}_I$ and the corresponding voxel features $V$ in each stage as depicted in Fig.~\ref{fig:DLGFM}, these features are initially fed into Siamese MLPs $\epsilon$. 
These MLPs are parameters shared, which aim to extract siamese representation features $F_I^s$ and $V^s$, respectively. The purpose of $\epsilon$ is to learn to extract homogeneous information between the two modalities. 
Then 
$F_I^s = [\mathbf{x}^I_1, \dots, \mathbf{x}^I_N]^\intercal$ 
and 
$V^s = [\mathbf{x}^V_1, \dots, \mathbf{x}^V_N]^\intercal$ 
are fed into the \textbf{Efficient Cross-Modal Interaction(ECMI)} to get cross-modal enhanced features, where $\mathbf{x}_i \in \mathbb{R}^{c}$ is a $c$ channel feature vector and $i$ indexes the spatial location of the feature vector. \\
Specifically, inspired by LatentGNN~\cite{LatentGNN} we introduce two sets of latent features $Z =[\mathbf{z}_1,\dots,\mathbf{z}_n]^\intercal$ for two modalities, respectively, where $z_i\in \mathbb{R}^{c}$ is a $c$-channel feature vector and the number of latent features $n \ll N$. Let $Z^I$ and $Z^V$ denote the latent feature for image modal and LiDAR modal, respectively. Then we build an augmented cross-modal graph $\mathcal{G}_L^c = (\mathcal{V}_L^c, \mathcal{E}_L^c)$ with $(2\times N) + (2\times n)$ nodes, $\mathcal{V}_L^c = \mathcal{V}^I \cup \mathcal{V}^I_h \cup \mathcal{V}^V \cup \mathcal{V}^V_h$, where $\mathcal{V}^I$ and $\mathcal{V}^V$ corresponds to the input features of image and LiDAR modal and $\mathcal{V}^I_h$ and $\mathcal{V}^V_h$ is associated with the latent features. We also define the graph connectivity by three subsets of the graph edges, $\mathcal{E}_v^I \cup \mathcal{E}_v^V \cup \mathcal{E}^c_h$, where $\mathcal{E}_v^I$ and $\mathcal{E}_v^V$ denote the connections between $\mathcal{V}^I$ and $\mathcal{V}^I_h$, $\mathcal{V}^V$ and $\mathcal{V}^V_h$, respectively. $\mathcal{E}_h^c$ are the graph edges within $\mathcal{V}^I_h \cup \mathcal{V}^V_h$.

\begin{figure}[t]
  \centering
  \includegraphics[width=\linewidth]{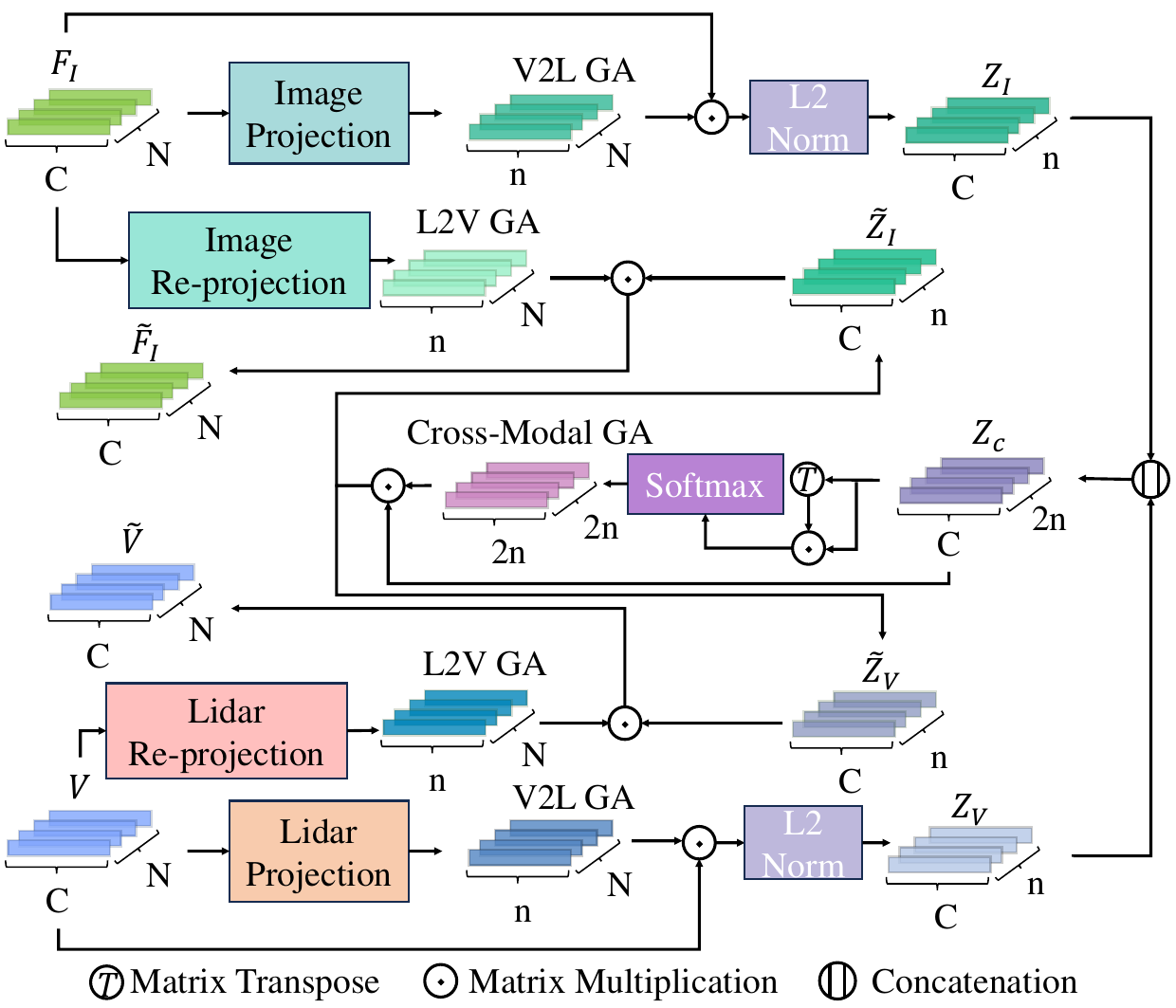}
  \caption{Structure of Efficient Cross-Modal Interaction.}
  \label{fig:ECI}
\end{figure}

Concretely, the \textbf{Efficient Cross-Modal Interaction} consists of three parts as shown in Fig.~\ref{fig:ECI}, where the GA represents Graph Affinity. 
First, it propagates messages from input feature nodes to latent nodes to obtain modal-specific latent features:
\begin{gather}
        \mathbf{z}^I_k = \sum_{i=1}^{N} \psi(x^I_i, \theta _k^{I})\mathbf{W}^\intercal x^I_i,~1 \le k \le n, \\
        \mathbf{z}^V_k = \sum_{i=1}^{N} \psi(x^V_i, \theta _k^{V})\mathbf{W}^\intercal x^V_i,~1 \le k \le n,
\end{gather}
where $\psi(x_i, \theta _k)$ encodes the affinity between the input feature node $x_i$ and the latent node $\mathbf{z}_k$, and $\theta_k$ is the parameter of the affinity function. 

To achieve non-local cross-modal integration while reducing computational costs and time complexity, we choose to do cross-attention in the latent space in a graph neural network manner. 
We first concatenate the two modal latent nodes to get cross-modal latent features $Z^c=[z^I_1, z^V_1, \dots, z^I_n, z^v_n]^\intercal$ and latent nodes $\mathcal{V}^c_h = \mathcal{V}^I_h \cup \mathcal{V}^V_h$,
then refine the latent features by propagating information in the fully connected cross-modal latent subgraph $(\mathcal{V}^c_h, \mathcal{E}^c_h)$:
\begin{equation}
    \tilde{\mathbf{z}}^c_k = \sum_{i=1}^{n}f(\sigma_k, \sigma_k, \mathbf{X})\mathbf{z}^c_i,~1 \le k \le n,
\end{equation}
where $f(\sigma_k, \sigma_i, \mathbf{X})$ represents the data-dependent pairwise relations between two latent nodes. $\sigma_k$ and $\sigma_i$ are the parameters for node $k$ and $i$, respectively. 

\begin{table*}[!t]
\centering
\setlength{\tabcolsep}{0.75mm}{
\begin{tabular}{l|c|cccccc|cccccc}
\hline
 & & \multicolumn{6}{c|}{KITTI \textit{test} set} & \multicolumn{6}{c}{KITTI \textit{val} set} \\ \cline{3-14} 
 & & \multicolumn{3}{c}{Car 3D AP} & \multicolumn{3}{c|}{Car BEV AP} & \multicolumn{3}{c}{Car 3D AP} & \multicolumn{3}{c}{Car BEV AP} \\
\multirow{-3}{*}{Method} & \multirow{-3}{*}{Modal.} & Easy & Mod. & Hard & Easy & Mod. & Hard & Easy & Mod. & Hard & Easy & Mod. & Hard \\ \hline
PointRCNN~\cite{Pointrcnn} & L & 87.0 & 75.6 & 70.7 & 92.1 & 87.4 & 82.7 & 88.9 & 78.6 & 77.4 & - & - & - \\
PV-RCNN~\cite{PVRCNN} & L & 90.3 & 81.4 & 76.8 & \textbf{95.0} & 90.7 & 86.1 & 92.6 & 84.8 & 82.7 & 95.8 & 91.1 & 88.9 \\
Voxel-RCNN$^{\dagger}$~\cite{VoxelRCNN} & L & 90.9 & 81.6 & 77.1 & 94.9 & 88.8 & 86.1 & 92.4 & 85.3 & 82.9 & 95.5 & 91.3 & 89.0 \\
M3DETR~\cite{M3DETR} & L & 90.3 & 81.7 & 77.0 & 94.4 & 90.4 & 86.0 & 92.3 & 85.4 & 82.9 & - & - & - \\
Octr~\cite{Octr} & L & 90.9 & 82.6 & 77.8 & 93.1 & 89.6 & 86.7 & 89.8 & 87.0 & 79.3 & - & - & - \\ 
\hline
MVXNet~\cite{MVXNet} & L+C & 83.2 & 72.7 & 65.2 & 89.2 & 85.9 & 78.1 & 85.5 & 73.3 & 67.4 & 89.5 & 84.9 & 79.0 \\
CLOCs~\cite{CLOCs} & L+C & 89.2 & 82.3 & 77.2 & 92.9 & 89.5 & 86.4 & 92.8 & 85.9 & 83.3 & 93.5 & 92.0 & 89.5 \\
Focals Conv~\cite{FocalConv} & L+C & 90.6 & 82.3 & 77.6 & 92.7 & 89.0 & 86.3 & - & - & - & - & - & - \\
VPFNet~\cite{VPFNet} & L+C & 91.0 & 83.2 & 78.2 & 93.9 & 90.5 & 86.3 & - & - & - & - & - & - \\
CAT-Det~\cite{CAT-Det} & L+C & 89.9 & 81.3 & 76.7 & 92.6 & 90.1 & 85.8 & 90.1 & 81.5 & 79.3 & - & - & - \\
HMFI~\cite{HMFI} & L+C & 88.9 & 81.9 & 77.3 & 93.0 & 89.2 & 86.4 & - & 85.1 & - & - & - & - \\
DVF~\cite{DenseFusion} & L+C & 89.4 & 82.5 & 77.6 & 93.1 & 89.4 & 86.5 & 93.1 & 85.8 & 83.1 & \textbf{96.2} & 91.7 &  89.2 \\
MLF-Det~\cite{MLF-Det} & L+C & 91.2 & 82.9 & 77.9 & 93.4 & 89.8 & 84.8 & 89.7 & \textbf{87.3} & 79.3 & - & - & - \\
GraphAlign~\cite{Graphalign} & L+C & 90.9 & 82.2 & 79.7 & 94.5 & 90.7 & 88.3 & 92.4 & 87.0 & 84.7 & 95.7 & \textbf{92.8} & 91.4 \\
SSLFusion(Ours) & L+C & \textbf{91.4} & \textbf{84.4} & \textbf{80.0} & 94.9 & \textbf{91.3} & \textbf{88.6} & \textbf{94.1} & 85.7 & \textbf{85.4} & 95.6 & 91.6 & \textbf{91.4} \\
\hline
\end{tabular}
}
\caption{Comparison of our model with state-of-the-art models on the KITTI test and validation sets for the car class. 'Mod.' indicates the moderate difficulty level, and $\dagger$ denotes the baseline method. The best results are highlighted in bold. L and C stand for the LiDAR and Camera modalities, respectively.}
\label{tab:kitti-test-val-car}
\end{table*}

Finally, we split the cross-modal interaction latent nodes $\tilde{\mathbf{Z}}^c$ back to two modal-specific latent nodes $\tilde{\mathbf{Z}}_I$ and $\tilde{\mathbf{Z}}_V$. We update the input features by propagating the messages from the latent nodes back to the feature nodes:
\begin{gather}
     \tilde{\mathbf{x}}^I_k = \alpha(\sum_{k=1}^{n} \psi(\mathbf{x}^I_i, \theta_k)\tilde{\mathbf{z}}_k),~1 \le i \le N, \\
      \tilde{\mathbf{x}}^V_k = \alpha(\sum_{k=1}^{n} \psi(\mathbf{x}^V_i, \theta_k)\tilde{\mathbf{z}}_k),~1 \le i \le N,
\end{gather}
where $\alpha$ is an activation function. The output of the final step is pulsed by the input feature as residual connection, in terms of output are $\tilde{F}_I =[\tilde{\mathbf{x}}^I_1 + \mathbf{x}^I_1, \dots, \tilde{\mathbf{x}}^I_N + \mathbf{x}^I_N]$ and $\tilde{V} =[\tilde{\mathbf{x}}^V_1 + \mathbf{x}^V_1, \dots, \tilde{\mathbf{x}}^V_N + \mathbf{x}^V_N]$. This process does not require explicitly maintaining the full affinity matrix and significantly reduces the computational cost. Our method only has linear complexity in terms of graph size, \ie, $O(N \cdot c \cdot n)$ and achieves $N/n$ times speed-up compared to QKV-based cross attention operations.

After getting the cross-modal interaction features $\tilde{F}^I$ and $\tilde{V}$ from \textit{Efficient Cross-Modal Interaction} aforementioned, we can get the final fused feature by:
\begin{gather}
    \begin{split}
            F_{f} = FuseConv(atten_I(F^s_I + V^s) \cdot \tilde{F}_I \\
            + atten_V(F^s_I + V^s) \cdot \tilde{V}),
    \end{split}
\end{gather}
where $atten_I$ and $atten_V$ are MLPs comprising two fully connected layers, a ReLU, and a Sigmoid. The learned attention weights from $atten_I$ and $atten_V$ effectively balance the contributions from image and point cloud features. The Fusion block contains a fully connected layer with a norm layer and a ReLU activation function.

\section{Experiments}
\subsection{Datasets and Metrics}
\textbf{KITTI Dataset.} 
The KITTI dataset stands as one of the most widely used multimodal 3D object detection datasets. 
It comprises 7481 training samples and 7518 testing samples consisting of RGB images and LiDAR point clouds, along with annotated 3D bounding boxes and object categories. 
Due to the unavailability of ground truth for the \textit{test} set, we follow common practice~\cite{PVRCNN, VoxelRCNN, MVXNet} and split the training samples into two sets: the train split set (referred to as the \textit{train} set) comprising 3712 samples and the validation split set (referred to as the \textit{val} set) comprising 3769 samples. 
When conducting the ablation studies, we train the model on the \textit{train} set and evaluate its performance on the \textit{val} set. 
For evaluating results on the \textit{test} set, we utilize the entire training samples, which include both the \textit{train} and \textit{val} sets. 
The results on the \textit{test} set are obtained by uploading to the server. We evaluate the performance using 3D and BEV Average Precision (AP) calculated from 40 recall positions. The Intersection over Union (IoU) thresholds for cars, cyclists, and pedestrians are set to 0.7, 0.5, and 0.5, respectively.

\noindent\textbf{DENSE Dataset.} 
DENSE Dataset serves as an object detection benchmark specifically designed to evaluate algorithms in challenging adverse weather conditions. 
Following the previous works~\cite{FogSim, GMIR}, we set a train-validation-test split ratio of 60\%, 15\%, and 25\%, respectively.  
Additionally, frames containing fewer than 3000 LiDAR points within the camera field of view are filtered out.
The train split contains all weather conditions, including clear day, clear night, dense fog, light fog and snow. 
We evaluate all models on two main conditions in the \textit{test} split, namely \textit{all weather} and \textit{dense fog}.
Specifically, the \textit{all weather} test set comprises 2998 scenes and the \textit{dense fog} test set comprises 88 scenes. 
Our experiments leverage the DENSE Dataset to evaluate the effectiveness of our proposed approach under diverse weather conditions.
As the same as the KITTI dataset, we choose the 3D and BEV AP calculated from 40 recall positions. Additionally, the IoU threshold for cars is set as 0.5.

\subsection{Implementation Details} 
Since the KITTI and DENSE datasets only provide the field of vision annotations and RGB images, we set the voxel size as (0.05, 0.05, 0.1) meters with the point cloud in the range of [0, 70.4], [-40.0, 40.0], and [-3.0, 1.0] meters along the X, Y, and Z axes, respectively.
And the image resolutions are set to 1224 $\times$ 370 pixels and 1920 $\times$ 1024, respectively.
For training on both datasets, ResNet50~\cite{ResNet} with traditional feature pyramid network (FPN~\cite{FPN}) is chosen as the image backbone. 
We employed the Voxel-RCNN~\cite{VoxelRCNN} head as the Region of Interest (ROI) head and a regular anchor head for the network.
Our model is trained from scratch in an end-to-end manner using the AdamW optimizer and the one-cycle policy with a learning rate of 0.01 with epoch 80. The batch size is set as 16 and 4 for KITTI and DENSE datasets, respectively.
\subsection{Results on KITTI Dataset}

\begin{table}[t]

\setlength{\tabcolsep}{0.6mm}
\begin{tabular}{c|c|ccc|ccc}
\hline
\multirow{2}{*}{Method} & \multirow{2}{*}{Modal.} & \multicolumn{3}{c|}{Cyc. 3D AP} & \multicolumn{3}{c}{Ped. 3D AP} \\
 &  & Easy & Mod. & Hard & Easy & Mod. & Hard \\ \hline
VoxelNet & L & 67.2 & 47.7 & 45.1 & 57.9 & 53.4 & 48.9 \\
SECOND & L & 78.5 & 56.7 & 52.8 & 58.0 & 51.9 & 47.1 \\
PointPillars & L & 82.3 & 59.3 & 55.3 & 58.4 & 51.4 & 45.2 \\
PointRCNN & L & 83.7 & 66.7 & 61.9 & 63.3 & 58.3 & 51.6 \\
Octr & L & 85.3 & 70.4 & 66.2 & 61.5 & 57.2 & 52.4 \\
Voxel-RCNN$^\dagger _*$ & L & 90.1 & 71.8 & 68.3 & 64.5 & 58.5 & 53.3 \\ \hline
Point-Fusion & L+C & 49.3 & 29.4 & 27.0 & 33.4 & 28.0 & 23.4 \\
MVXNet & L+C & 73.1 & 54.6 & 51.4 & 61.2 & 55.7 & 52.2 \\
CLOCs & L+C & 90.3 & 64.8 & 59.6 & 60.3 & 54.2 & 46.4 \\
SSLFusion & L+C & \textbf{93.0} & \textbf{74.3} & \textbf{71.6} & \textbf{66.0} & \textbf{60.4} & \textbf{56.2} \\
\hline
\end{tabular}
\caption{
Performance comparison on KITTI val set for Cyclist and Pedestrian classes,
$\dagger$ means baseline models and $*$ means our re-implementation results. Cyc. and Ped. denote the Cyclist and Pedestrian classes respectively.}
\label{tab:kitti-val-cyc-ped}
\end{table}

\noindent\textbf{Comparison on \textit{test} set.}
Here, we evaluate the performance on the KITTI \textit{test} set for car class. 
We compare our proposed SSLFusion with other state-of-the-art methods. The results are reported in Table~\ref{tab:kitti-test-val-car}. 
It can be observed that SSLFusion outperforms the state-of-the-art methods in terms of 3D AP across all levels. 
Additionally, it achieves superior performance in terms of BEV AP on both moderate and difficult levels.
Compared with the baseline method Voxel-RCNN~\cite{VoxelRCNN}, 
our SSLFusion increases the 3D AP by 2.76\% and 2.98\% on moderate and hard difficulty levels, respectively. Additionally, it enhances the BEV AP by 2.43\% and 2.42\% on moderate and hard difficulty levels, respectively.
Among existing multimodal 3D object detection methods, the recently proposed VPFNet~\cite{VPFNet} reports the best results on the moderate level. Compared with VPFNet, our SSLFusion achieves improvements of 1.17\% on the moderate level. 

\noindent\textbf{Comparison on \textit{val} set.}
Here, our SSLFusion is compared with recent state-of-the-art methods on the KITTI \textit{val} set specifically for the car class. Table~\ref{tab:kitti-test-val-car} gives the results.
Our SSLFusion obtains the best results on the easy and hard levels in 3D AP.
Our SSLFusion outperforms Voxel-RCNN with an absolute gain of 1.73\%, 0.42\%, and 2.50\% on the easy, moderate, and difficult levels, respectively, in terms of 3D AP.
It can be observed that among the three difficulty levels, the most substantial improvement occurred in the difficult level, indicating employing our method to aggregate image features into point cloud features can effectively enhance the detection accuracy of hard samples.
In addition, we also report the performance on the KITTI \textit{val} set for cyclist and pedestrian classes.
As shown in Table~\ref{tab:kitti-val-cyc-ped}.
It can be observed that our method obtain the best performance on cyclist and pedestrian classed on the KITTI \textit{val} set across all difficulty levels in both 3D AP and BEV AP. 
The results for cyclist and pedestrian classes demonstrate the robustness and effectiveness of our SSLFusion.
Additionally, we test the inference time of SSLFusion on an NVIDIA RTX 3090 GPU, it achieves around 11.3 frames per second (FPS).

\begin{table}[t]
\setlength{\tabcolsep}{1mm}{
\begin{tabular}{c|c|ccc|ccc}
\hline
 & & \multicolumn{3}{c|}{3D AP} & \multicolumn{3}{c}{BEV AP} \\
\multirow{-2}{*}{\textit{W}} & \multirow{-2}{*}{Method} & Easy & Mod. & Hard & Easy & Mod. & Hard \\ \hline
 & SECOND & 65.7 & 63.1 & 59.1 & 69.9 & 68.2 & 64.6  \\
 & PartA$^2$ & 64.7 & 61.6 & 56.7 & 69.1 & 67.2 & 62.6  \\
 & MVXNet & 71.8 & 70.4 & 65.8 & 75.7 & 75.3& 71.2 \\
 & Voxel-RCNN$^{\dagger}$ & 72.0 & 69.7 & 67.0 & 73.3 & 72.0 & 69.4  \\
 & VoxelNeXt & 73.2 & 71.6 & 68.7 & 77.0 & 75.7 & 72.9  \\
\multirow{-7}{*}{\textit{AW}} & SSLFusion & \textbf{81.4} & \textbf{79.0} & \textbf{74.0} & \textbf{84.7} & \textbf{82.3} & \textbf{77.3} \\ \hline
 & SECOND & 56.3 & 58.8 & 55.0 & 63.2 & 64.7 & 62.2  \\
 & PartA$^2$ & 50.6 & 51.0 & 46.7 & 57.7 & 58.9 & 56.1  \\
 & MVXNet& 60.0 & 60.9 & 58.1 & 63.8 & 66.2 & 63.3  \\
 & Voxel-RCNN$^{\dagger}$ & 69.0 & 69.3 & 66.2 & 72.2 & 73.2 & 69.8  \\
 & VoxelNeXt & 66.0 & 66.5 & 63.1 & 71.0 & 72.0 & 69.0 \\
\multirow{-7}{*}{\textit{DF}} & SSLFusion & \textbf{78.2} & \textbf{76.4} & \textbf{71.8} & \textbf{82.4} & \textbf{80.5} & \textbf{77.9} \\
\hline
\end{tabular}
}
\caption{
Performance comparison on the DENSE test set for the Car class, including all-weather and dense fog splits.
$\dagger$ means baseline models. W, AW and DF refer to weather split, all-weather set and dense fog set, respectively. Best in bold.}
\label{tab:dense-car-test}
\end{table}

\subsection{Results on DENSE Dataset}
To evaluate generalization and weather robustness, we conduct experiments on the DENSE dataset, which comprises various weather conditions.
Different from KITTI, the DENSE dataset lacks an official leaderboard server for comparing the detection performance of existing methods. Therefore, for fair comparisons, we train some recent 3D object detectors, including SECOND~\cite{SECOND}, Part-A$^2$~\cite{PartA2},
Voxel-RCNN, VoxelNeXt~\cite{Voxelnext} and MVXNet, using the same dataset and the code provided by MMDetection3D~\cite{mmdet3d} and OpenPCDet~\cite{openpcdet}.
The results for both the \textit{all weather} and \textit{dense fog} sets are presented in Table~\ref{tab:dense-car-test}.
It can be observed that SSLFusion outperforms state-of-the-art methods in terms of 3D and BEV AP across all difficulty levels in both the all-weather and dense fog sets. 
Among existing methods, VoxelNeXt reports the best results on the \textit{all weather} set with 3D AP of 73.17\%, 71.64\%, and 68.73\% and BEV AP of 76.94\%, 75.66\%, and 72.93\% across all difficulty levels. Our SSLFusion method outperforms VoxelNeXt, achieving an absolute gain of 8.22\%, 7.39\%, and 5.27\% in terms of 3D AP across difficulty levels, as well as a gain of 7.75\%, 6.65\%, 4.4\% in terms of BEV AP across difficulty levels.
Additionally, on the \textit{dense fog} set, the Voxel-RCNN reports the best results among existing methods with 3D AP of 69.03\%, 69.31\%, 66.24\% and BEV AP of 72.23\%, 73.18\%, 69.79\% across all difficulty levels. Our SSLFusion achieves superior results with 3D AP of 78.15\%, 76.36\%, 71.76\% and BEV AP of 82.42\%, 80.47\%, 77.90\% across all difficulty levels.
This observation highlights the superior detection accuracy achieved by our method under various complex weather conditions in the real world, demonstrating the robustness and generalization capability of our approach.

\subsection{Ablation Study}

\begin{table}
\setlength{\tabcolsep}{1.8mm}{
\begin{tabular}{cccc|ccc}
\hline
\multirow{2}{*}{Method} & \multirow{2}{*}{SAF} & \multirow{2}{*}{SAM} & \multirow{2}{*}{LFM} & \multicolumn{3}{c}{3D AP} \\ \cline{5-7} 
 &  &  &  & \multicolumn{1}{c|}{Easy} & \multicolumn{1}{c|}{Mod.} & Hard \\ \hline
Baseline & - & - & - & \multicolumn{1}{c|}{92.38} & \multicolumn{1}{c|}{85.29} & 82.86 \\ \hline
\multirow{3}{*}{Ours} & \checkmark & - & - & \multicolumn{1}{c|}{92.93} & \multicolumn{1}{c|}{85.50} & 83.01 \\
 & \checkmark & \checkmark & - & \multicolumn{1}{c|}{93.91} & \multicolumn{1}{c|}{85.67} & 84.75 \\
 & \checkmark & \checkmark & \checkmark & \multicolumn{1}{c|}{94.12} & \multicolumn{1}{c|}{85.71} & 85.36 \\ \hline
\multicolumn{4}{c|}{Improvements} & \multicolumn{1}{c|}{+1.74} & \multicolumn{1}{c|}{+0.42} & +2.50 \\
\hline
\end{tabular}%
}
\caption{Effect of each component of our SSLFusion on KITTI val set for the car class.}
\label{tab:ablation}
\end{table}

In this section, we perform an ablation study to analyze the impact of each component of our SSLFusion, including the Scale-Aligned Fusion (SAF), 3D-to-2D Space Alignment Module (SAM), and Latent Cross-Modal Fusion Module (LFM), on the KITTI \textit{val} set for the car class. 
The Voxel RCNN is chosen as the baseline method.
The results are given in Table~\ref{tab:ablation}.
It can be observed that compared with the baseline method, our SSLFusion obtains an absolute gain of 1.55\% of average 3D AP over three difficulty levels.

\noindent\textbf{Effect of Scale-Aligned Fusion Strategy (SAF).} 
From Table~\ref{tab:ablation}, it is evident that SAF leads to performance gains of 0.55\%, 0.21\%, and 0.15\% in terms of 3D AP on the easy, moderate, and hard difficulty levels, respectively. 
The performance improvements can be attributed to SAF significantly reducing irrelevant information during fusion while minimizing disparities between the two modalities. This is achieved by aggregating features extracted from images and LiDAR point clouds in a scale-aligned manner, provides stronger hierarchical information for scale-aligned fused features and significantly reduces information loss during downsampling through its substantial residual connections.

\noindent\textbf{Effect of 3D-to-2D Space Alignment Module (SAM).} 
From Table~\ref{tab:ablation}, we can observe that the 3D-to-2D Space Alignment Module (SAM) can bring 0.98\%, 0.17\%, and 1.74\% performance gains in terms of 3D AP on the easy, moderate, and hard difficulty level, respectively.
The most significant gains are observed on hard difficulty levels, indicating that SAM is particularly effective at handling challenging examples. This is mainly attributed to the fact that SAM incorporates 3D information from the 3D voxels with image features to achieve 3D-to-2D space alignment. 

\noindent\textbf{Effect of Latent Cross-Modal Fusion Module (LFM).} 
From Table~\ref{tab:ablation}, it can be observed that LFM enhances the 3D AP by 0.21\%, 0.04\%, and 0.61\% on the easy, moderate, and hard difficulty levels, respectively.
The performance gain highlights the significance of our LFM, which enriches the non-local context information during feature fusion between the two modalities. This enrichment is useful in inferring the existence and locations of objects.
Additionally, compared to the QKV-based attention fusion method on single-stage fusion, our latent cross-modal fusion achieves better accuracy (86.68\% vs. 86.49\%) and faster inference (10.75 FPS vs. 9.69 FPS), further demonstrating its efficiency.

\section{Conclusion}
In this work, we introduce SSLFusion, a novel approach for multimodal 3D object detection that focuses on aligning the scale and spatial discrepancies between features from 2D images and 3D point clouds. 
Our SSLFusion leverages a multi-stage fusion strategy that integrates features from both modalities, ensuring a more coherent and discriminative representation. 
A key contribution of SSLFusion is the Scale-Aligned Fusion Strategy, which mitigates scale misalignment by fusing two modalities features at each stage of the detection pipeline. 
The 3D-to-2D Space Alignment module reduces the inter-modal gap by attributing 3D spatial information to 2D features.
Additionally, the Latent Cross-Modal Fusion module efficiently and effectively captures non-local contexts by introducing latent space, which significantly reduces computation time and improves detection accuracy.
Experiments conducted on the KITTI and DENSE datasets demonstrate SSLFusion's superior performance. It outperforms existing state-of-the-art methods and showcases its robustness under various conditions and object scales. 

\section{Acknowledgements}

This research was funded by the National Key Research and Development Program of China (Grant No. 2022ZD0160400), the National Natural Science Foundation of China (Grant No. 62206031, 62301092, and 62271346).

\bibliography{aaai25}

\end{document}